\documentclass[10pt]{article}
\usepackage{epi-report}
\usepackage{placeins}
\usepackage{needspace,float}
\usepackage{fontawesome5}
\hypersetup{pdftitle={Épi: Towards a Persistent, Self-Improving Agent for Autonomous Research},pdfauthor={Yue Huang, Zhangchen Xu, Xiaonan Luo, Zheyuan Liu, Lichi Li, Yiping Wang, Peiyang Song, Eric Gabriel Ramos, Enze Xie, Olexandr Isayev, Paul Liang, Nouha Dziri, Huan Sun, Jianfeng Gao, Ying Liu, Misha Sra, Xiangliang Zhang, Alex Pentland, Zichen Chen},pdfsubject={Autonomous research agents, persistent research, verifiable optimization, and computer-assisted proofs}}
\newcommand{\epi}{Épi}

\newtheorem{theorem}{Theorem}
\title{Épi: Towards a Persistent, Self-Improving Agent\\for Autonomous Research}
\author{%
  Yue Huang\textsuperscript{1,2}\quad
  Zhangchen Xu\textsuperscript{1,3}\quad
  Xiaonan Luo\textsuperscript{1,2}\quad
  Zheyuan Liu\textsuperscript{1,2}\quad
  Lichi Li\textsuperscript{1}\\[2pt]
  Yiping Wang\textsuperscript{3}\quad
  Peiyang Song\textsuperscript{7}\quad
  Eric Gabriel Ramos\textsuperscript{4}\quad
  Enze Xie\textsuperscript{6}\quad
  Olexandr Isayev\textsuperscript{7}\\[2pt]
  Paul Liang\textsuperscript{8}\quad
  Nouha Dziri\textsuperscript{9}\quad
  Huan Sun\textsuperscript{10}\quad
  Jianfeng Gao\textsuperscript{11}\quad
  Ying Liu\textsuperscript{5}\quad
  Misha Sra\textsuperscript{12}\\[2pt]
  Xiangliang Zhang\textsuperscript{2}\quad
  Alex Pentland\textsuperscript{8,13}\quad
  Zichen Chen\textsuperscript{1,12,13}}
\affiliations{%
  \affiliationentry{1}{bake-ai}{Bake AI}\quad
  \affiliationentry{2}{notre-dame}{University of Notre Dame}\quad
  \affiliationentry{3}{washington}{University of Washington}\quad
  \affiliationentry{4}{stevens}{Stevens Institute of Technology}\\[4pt]
  \affiliationentry{5}{scale}{Scale AI}\quad
  \affiliationentry{6}{nvidia}{NVIDIA}\quad
  \affiliationentry{7}{cmu}{Carnegie Mellon University}\quad
  \affiliationentry{8}{mit}{MIT}\quad
  \affiliationentry{9}{cohere}{Cohere Labs}\\[4pt]
  \affiliationentry{10}{ohio-state}{The Ohio State University}\quad
  \affiliationentry{11}{microsoft}{Microsoft}\quad
  \affiliationentry{12}{ucsb}{UC Santa Barbara}\quad
  \affiliationentry{13}{stanford}{Stanford University}}
\date{Technical Report \quad October 2026}
\begin{document}
\maketitle
\begin{abstract}
We present Épi, an autonomous research agent that sustains investigations across model sessions and evolves its methods through accumulated experience. Épi combines persistent research history, hierarchical memory, branching and independent review to maintain research continuity, learn from experiments, and explore and revisit alternative approaches. Its experiments run across session boundaries, and a fixed verifier evaluates their outputs. Épi produced 120 record-setting results, comprising 20 numerical bound records across 12 families and 100 construction records across eight families. These include an approximately 84-fold narrower interval for the $n$-queens constant and a more than 3.6-million-fold narrower rigorous interval for the quartic Feigenbaum attractor dimension. Exact dual certificates establish that the Cohn--Elkies linear-programming bound cannot be sharp in dimensions 10 and 11. Beyond numerical improvements, Épi gives a computer-assisted proof of Tuza's conjecture for split graphs with an eight-vertex clique part and at most three triangle-active neighborhood types, with arbitrary multiplicities. We extend this research process to algorithmic coding with Épi-turbo, a lightweight configuration that develops executable solvers for tasks including vehicle routing, scheduling and production planning. On 30 FrontierOR tasks, Épi-turbo achieves a 5.4\% higher total benchmark score than Codex at 64\% of its model cost, using the same underlying model and reasoning effort. Épi is designed to support human understanding by connecting computational results to explicit methods and mathematical arguments that researchers can examine and build upon.

\par\vspace{8pt}
{\centering\fontsize{10.5}{13}\selectfont\sffamily\bfseries
\href{https://bakeai.inc/research/articles/epi/}{\faGlobe\hspace{0.45em}Blog}\hspace{2em}
\href{https://github.com/BakeLab/Epi-results}{\faGithub\hspace{0.45em}Results on GitHub}\par}
\end{abstract}

\section{Introduction}
Autonomous research agents must conduct experiments, interpret results and revise their methods over many interactions. FunSearch and AlphaEvolve demonstrate how agents can combine code generation with executable evaluation to discover new programs and constructions \citep{funsearch,alphaevolve}. The AI Scientist-v2 extends automation to an experimental workflow \citep{ai-scientist}. When an investigation requires more work than one model session can support, however, the agent must manage three additional challenges.

\paragraph{First, maintaining research continuity}
Experiments and research decisions span multiple model sessions. To continue an investigation, the agent needs access to running jobs, code, intermediate results and reasons for earlier decisions. Kosmos maintains a shared research state, InternAgent-1.5 organizes long-horizon memory, and ScienceFlow archives recoverable execution states \citep{kosmos,internagent,scienceflow}. These mechanisms preserve information across steps. As the history grows, the agent must also select what a bounded model context needs: the current experimental state, relevant lessons and reasons an earlier approach failed. This need motivates hierarchical memory that separates immediate context from reusable experience.

\paragraph{Second, learning from experimental experience}
An agent needs to turn experimental observations into knowledge that guides subsequent decisions. This requires identifying which choices produced a result, under what conditions the method applies and which evidence supports the lesson. The agent must retain this knowledge and retrieve it for relevant tasks. MemoHarness addresses experience reuse by building a memory bank during training and retrieving execution diagnoses and reusable patterns to adapt the harness at test time \citep{memoharness}. In an ongoing research investigation, the agent must also update its memory as new experiments arrive, linking each lesson to the conditions and evidence that support it.

\paragraph{Third, exploring alternatives and backtracking}
A single research direction explores only part of the available methods. The AI Scientist-v2 uses agentic tree search to develop experimental alternatives, and ScienceFlow can resume from archived executable states \citep{ai-scientist,scienceflow}. These systems demonstrate the value of branching and backtracking. For research with checkable outputs, the agent must preserve each alternative's implementation and evidence, compare candidates under a common evaluation rule and recover earlier work when new findings make it useful.

To address these challenges, we introduce \textbf{Épi}, a persistent and evolving autonomous research agent. Épi uses hierarchical memory to resume investigations and retrieve experience from earlier work. It maintains versioned branches to develop alternative methods and backtrack to saved implementations, and incorporates independent review when assessing evidence and choosing its next experiment. The agent operates in verifiable research environments that evolve between \emph{campaigns}, where a campaign is a complete investigation that may span multiple model sessions. Within a campaign, every branch uses the same feasibility checks, objective and comparison value. A verified finding can then establish a new reference or motivate a new task for the next campaign. Persistent execution and storage carry the research state across sessions.

Mathematical discovery provides a concrete demonstration of this research process: the agent develops methods whose outputs can be checked as constructions, bounds or proofs. We also evaluate its applicability to algorithmic coding, where the agent must design and implement a solver that handles new problem instances efficiently. Our lightweight configuration, Épi-turbo, achieves a 5.4\% higher total benchmark score than Codex on 30 FrontierOR tasks at 64\% of its model cost, using the same underlying model and reasoning effort (Section~\ref{sec:turbo}). These tasks include routing, scheduling, production planning and graph optimization.

Our contributions are an autonomous research agent that integrates hierarchical memory, branching and independent review to sustain and refine its investigations; a verification protocol that evaluates changing research methods under a fixed ruler and gain floor; and a collection of 120 record-setting results, comprising 100 improved constructions and 20 numerical bounds. Each record improves a documented literature or registry reference for the same problem and normalization; we count one final result per instance or parameter setting, as detailed in Section~\ref{sec:evaluation}. These results include an approximately 84-fold narrower interval for the $n$-queens constant and a more than 3.6-million-fold narrower rigorous interval for the quartic Feigenbaum attractor dimension. Exact certificates establish that the Cohn--Elkies bound is not sharp in dimensions 10 and 11. Épi also proves Tuza's conjecture for split graphs with an eight-vertex clique part and at most three triangle-active neighborhood types, with arbitrary multiplicities. We release the constructions and supporting certificates \citep{epi-results}.

Our goal is to connect self-improvement with human understanding: researchers should be able to inspect how a method evolves, explain why a result holds, and use that knowledge to guide further work. Épi retains implementations, experimental artifacts, and research notes across branches and sessions, allowing researchers to inspect the development of a method and assess its supporting evidence. Its constructions, certificates, and mathematical arguments further connect computational outputs to findings that researchers can examine and build upon (Section~\ref{sec:human-understanding}).

\Needspace{8\baselineskip}
\section{Research Setting}
\label{sec:setting}
\subsection{Problems and Evidence}
We study research tasks with a reference result, an explicit budget and outputs that can be checked. A curator supplies the question and source material; the agent chooses the methods and experiments. To compare the agent's output with the reference, we specify an evaluation rule, called a \emph{ruler}. The ruler defines the submission format, feasibility conditions, objective, normalization and comparison value.\footnote{A reported benchmark score alone does not specify feasibility conventions, normalization or input validation. These choices determine whether two values measure the same quantity on valid objects. We introduce the ruler to make them explicit before search.}

The agent submits the information required to check its claim. For a packing task, it supplies coordinates and orientations. For an optimization bound, it may supply a dual vector or another certificate. For a proof task, it supplies witnesses and an argument explaining why the checked objects establish the stated claim.

For a minimization problem, we write $V(x)\in\{0,1\}$ for the verifier's feasibility decision, $f(x)$ for the objective and $b$ for the reference value. We define the improvement as $\Delta(x)=b-f(x)$. We also specify a positive \emph{gain floor} $\delta>0$, the smallest improvement that the task counts as meaningful.\footnote{Counting every positive score change as progress can include differences due to reporting precision or routine post-processing. We derive the floor from the objective's granularity and known transformations, and record its rationale with the task. For an integer objective with unit granularity, a floor of one admits a one-unit improvement.}
Épi accepts an improvement when
\begin{equation}
 V(x)=1,\qquad \Delta(x)\geq\delta.
 \label{eq:gain}
\end{equation}
For maximization, we instead define $\Delta(x)=f(x)-b$. The positive floor excludes ties and admits an improvement exactly equal to the threshold. If a task imposes no minimum gain ($\delta=0$), we instead require $\Delta(x)>0$. We fix the feasibility tolerances and gain floor before the investigation. The verifier uses the tolerances to check constraints, while Épi uses the floor to decide whether the improvement counts. For a rigorous numerical bound, we additionally require exact arithmetic or outward rounding that controls the numerical error.

We establish a new record by verifying feasibility and improvement over the best established result for the same problem and normalization. Each final artifact retains its comparison value, source and any rounding convention. For a theorem, the proof connects the verified objects to the full stated claim.

\subsection{Campaigns and Sessions}
We call a complete investigation a \emph{campaign}. A campaign may contain several agent sessions and concurrent experiments. Each session gives the agent a bounded context in which to inspect evidence, modify code and choose its next action. Épi stores the research state outside that context and keeps experiments running across session boundaries. A later session retrieves the saved state and resumes the investigation.

The curator may provide a published construction as the initial reference. The agent can consult the literature, implement alternative methods and select experiments within the campaign's budget. We keep model weights fixed in the campaigns reported here. The agent adapts its methods through experimental feedback, retrieved memory and independent review.

\Needspace{8\baselineskip}
\section{The Épi Research Loop}
\label{sec:method}
\begin{figure}[t]
\centering
\includegraphics[width=\linewidth]{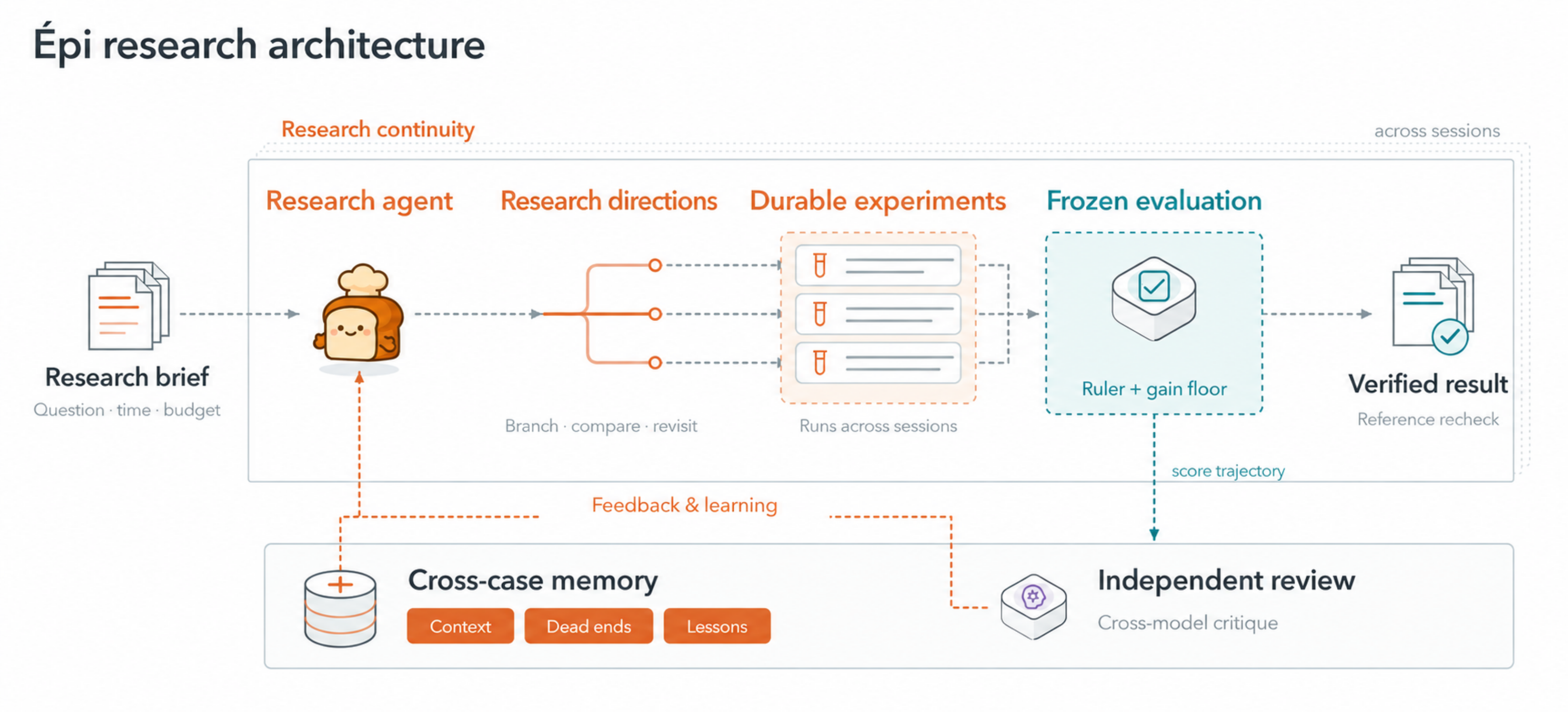}
\caption{Épi's research architecture. The agent develops methods on versioned branches and retrieves saved history and cross-case lessons. An independent model reviews the evidence. The environment checks each candidate under a fixed ruler and gain floor. The execution infrastructure supports experiments, persistent storage and isolated verifier execution across sessions. Section~\ref{sec:setting} defines the acceptance rule.}
\label{fig:architecture}
\end{figure}

Figure~\ref{fig:architecture} shows Épi's research loop. The agent conducts experiments in verifiable environments and improves its methods through branching, hierarchical memory and independent review. We describe the evolving environments and the mechanisms supporting the agent's self-improvement below.

\subsection{Evolving Research Environments}
Comparing research methods requires a common objective and consistent validity checks. We establish these before search by checking that the scorer reproduces the reference result, testing input validation and fixing the configuration, feasibility checks and gain criterion. The verifier then evaluates every branch under this shared rule. This gives Épi a consistent basis for comparing methods throughout the investigation.

We also specify the evidence needed for each type of claim. The verifier checks geometric constraints for a construction. For a rigorous bound, we require a certificate with exact arithmetic or controlled numerical error. For a computer-assisted theorem, we require both the finite checks and a mathematical reduction showing why those checks establish the full statement. The agent must develop this reduction or certificate when the task requires one.

Épi's research environments evolve with verified findings. Improved results provide new reference values for subsequent investigations, while new research questions motivate changes to task formulations, representations and verification requirements. We incorporate these changes when defining the next campaign. For example, a numerical estimate can lead to a certification task whose environment includes certificate representations and rigorous error control. Each campaign uses a fixed evaluation rule to compare methods, and successive campaigns build on the verified findings of earlier work.

\subsection{Self-improving Research Agents}
Épi improves its research methods through accumulated experience. The agent develops and compares alternatives on versioned branches, retains and retrieves experience through hierarchical memory, and uses independent review to reassess its research decisions. Together, these mechanisms allow Épi to refine its approach across experiments and campaigns.

\paragraph{Branches and backtracking}
A single approach restricts which formulations and search regions the agent explores. Épi maintains branches to develop alternatives while preserving each method's code and evidence. Each branch stores an implementation, tests and notes in a versioned repository. The agent saves changes as commits and can create a new branch from an earlier commit. It compares the resulting experiments under the same ruler.

The agent backtracks by returning to a saved implementation and trying a different continuation. It can also revisit an existing branch and extend the work already recorded there. When the agent abandons a branch, it records the reason and retains the branch for later inspection. For a packing problem, the agent might refine coordinates on one branch and change the contact pattern on another. It can compare both methods before allocating more compute.

\paragraph{Hierarchical memory}
Drawing on work on human and agent memory \citep{atkinson-shiffrin,memgpt,generative-agents}, Épi uses hierarchical memory to accumulate and reuse research experience. Short-term memory supports the current investigation, while long-term memory retains experience for subsequent research. As illustrated in Figure~\ref{fig:architecture}, memory records include research context, lessons that capture reusable findings, and dead ends that document unproductive approaches and why they were abandoned. The agent retrieves these records to guide its next experiment and updates them as new evidence becomes available.

Épi stores code, artifacts and detailed research notes in a persistent filesystem and keeps retrievable lessons and dead ends in a memory store outside the model context. At a session boundary, the agent records its findings and a handover note identifying the current state and next steps. A subsequent session receives relevant memory records and can open the referenced files as needed. This supplies a bounded working context without replaying the full history of earlier sessions.

\paragraph{Independent review}
The agent that develops a method also chooses which evidence to examine, giving its assessment a limited perspective. Prior work on self-correction motivates adding feedback from outside the agent's own reasoning process \citep{huang-self-correction}; research on reward hacking further motivates checking whether a score increase reflects the intended task \citep{denison-reward-tampering}. Épi combines a separate review of research progress with artifact verification under the fixed ruler.

Épi supplies a review model with the score trajectory, attempted approaches, retrieved findings, comparison baseline and resource use. The reviewer assesses what the evidence supports and whether the agent's interpretation of progress is justified. Épi records this assessment for subsequent sessions, and the agent uses it when selecting further experiments or revisiting a branch. We use a different model for an independent review and record the reviewer's identity. The reviewer evaluates the research direction; the isolated verifier checks the candidate's validity and gain.

\paragraph{Research transparency}
Épi records alternative implementations, their experimental outcomes, and the assessments used to guide subsequent work. These records allow researchers to compare branches, inspect the evidence associated with a methodological change, and revisit the implementation on which a conclusion depends. Lessons summarize reusable findings, while references to code and experimental artifacts let researchers examine the evidence behind those summaries.

\paragraph{Execution support}
We use Breadboard as Épi's execution infrastructure. Breadboard provides isolated experimental environments, supports long-running jobs across model sessions and stores research history, including code, branch states and experimental outputs. It also runs the fixed verifier and preserves the state needed to recover and continue an investigation. The agent uses these services to execute experiments, inspect results and resume its research across sessions.

\FloatBarrier

\section{Experimental Results}
\label{sec:evaluation}
\subsection{Setup and Counting}
We ran Épi over a few days using OpenAI frontier models (GPT-5.6 Sol and GPT-6-Astra). Computational budgets were set per problem. Published methods and constructions supplied starting points for several campaigns, including the packing case in Section~\ref{sec:packing}. Each comparison uses the same task and evaluation convention for the reference and final artifact. All record comparisons are based on literature and registry references available as of September 16, 2026.

The current collection contains \textbf{120 record-setting results}: \textbf{20 numerical bound records across 12 families} and \textbf{100 construction records across eight families}. A separate Tuza theorem extension brings the report to 21 families in total and is excluded from the record count.

We count one final result per distinct instance or parameter setting. An enclosing interval counts once, and intermediate improvements do not add records. Figure~\ref{fig:families} gives the family breakdown.

The results repository retains the final artifacts and comparison values \citep{epi-results}. Some evolving registries already list the new results. Our tables retain the earlier values used for comparison so the improvement remains traceable after those updates.

\subsection{Construction Records}
\begin{figure}[H]
\centering\includegraphics[width=0.78\linewidth]{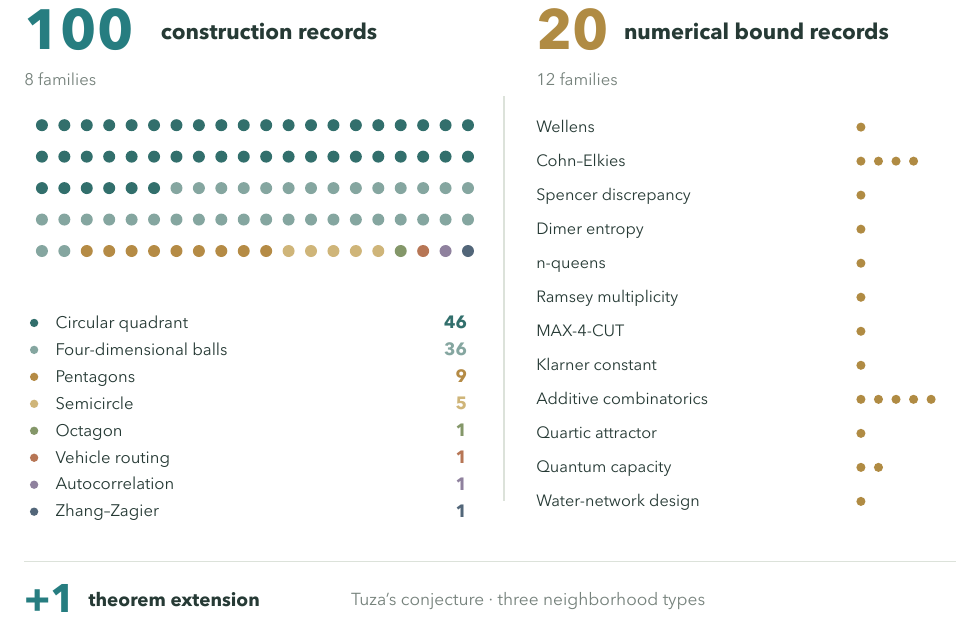}
\caption{The 120 records, grouped by collection and family. Each dot is one final record. Multiple parameter settings within a family count separately; repeated improvements to the same setting count once. The Tuza theorem extension is shown separately.}
\label{fig:families}
\end{figure}

Of the 100 construction records, 97 concern geometric packing. The remaining three improve vehicle routing, the signed-maximum variant of the third autocorrelation inequality and a Zhang--Zagier construction. Figure~\ref{fig:families} gives the counts for all eight families. Packing references come from Packomania and Erich Friedman's Packing Center \citep{packomania,friedman}. Appendix~\ref{app:construction} defines the objectives and gives these three comparisons.

\subsection{Numerical Bound Records}
Table~\ref{tab:mathsummary} summarizes the 12 numerical families and how their bounds are verified. Appendix~\ref{app:math} states the claims with their precise values and supporting witnesses.

\begin{table}[H]
\caption{Twenty numerical records across 12 families. Arrows indicate the improving direction. Interval widths are reduced. This overview uses rounded decimals. Precise bounds appear in the appendix and Table~\ref{tab:ce}.}
\label{tab:mathsummary}
\small
\renewcommand{\arraystretch}{1.2}
\begin{tabularx}{\linewidth}{>{\raggedright\arraybackslash}m{3.10cm} >{\raggedleft\arraybackslash}m{2.80cm} >{\raggedleft\arraybackslash}m{2.70cm} L}
\toprule\headrow
\textbf{Quantity} & \textbf{Previous bound} & \textbf{Épi} & \textbf{Evidence}\\\midrule
Wellens' $D_\infty$ $\downarrow$ & $4.393500$ & $4.365467$ & Rational witnesses\\
CE dual bounds $\uparrow$ & \shortstack[r]{$1/20,\;1/24,$\\$1/24,\;1/28$} & \shortstack[r]{Table~\ref{tab:ce}:\\$d=9,10,11,13$} & Exact dual certificates\\
Spencer constant $\uparrow$ & $7/\sqrt{17}$ & $5/\sqrt8$ & All 256 sign assignments\\
Dimer entropy $\downarrow$ & $0.452130$ & $0.452010$ & Integer transfer inequalities\\
$n$-queens interval width & $3.30\!\times\!10^{-7}$ & $3.94\!\times\!10^{-9}$ & Rational and interval checks\\
Ramsey $c_{4,5}$ $\downarrow$ & $0.007932$ & $0.007865$ & A 1,024-vertex graph\\
MAX-4-CUT ratio $\downarrow$ & $0.987085$ & $0.986411$ & Integer weighted gadget\\
Klarner $\lambda$ $\downarrow$ & $4.52380$ & $4.29569$ & Recurrence certificate\\
$B_h[g]$ coefficients $\downarrow$ & Timmons' Table~1 & Five bounds, $h=3,\ldots,7$ & Interval-certified kernels\\
Quartic dimension interval width & $7.80\!\times\!10^{-3}$ & $2.13\!\times\!10^{-9}$ & Full-domain interval proof\\
\shortstack[l]{Quantum noise $\uparrow$\\(XZ / Two-Pauli)} & \shortstack[r]{$0.118371$\\$0.118067$} & \shortstack[r]{$0.1198803$\\$0.1193415$} & Positive coherent information\\
Waterund36 lower bound $\uparrow$ & $655.544171$ & $662.807028$ & Exact bound certificate\\\bottomrule
\end{tabularx}
\end{table}

\FloatBarrier
\section{Case Studies}
\label{sec:cases}
\subsection{Refining a Published Packing}
\label{sec:packing}
The task is to place 34 non-overlapping regular pentagons of unit side length inside an equilateral triangle of minimum side length. Each pentagon has a center and orientation. The evaluator checks containment and overlap and measures the required triangle size.

Épi starts from a published arrangement with triangle side length approximately $12.9814$ \citep{friedman}. It jointly refines the pentagons' centers and orientations, checking containment and overlap after each candidate update. The final side length is approximately $12.9756$, a reduction of $0.0445\%$. The released coordinates specify the construction at full precision.

\begin{figure}[htb]
\centering\includegraphics[width=\linewidth]{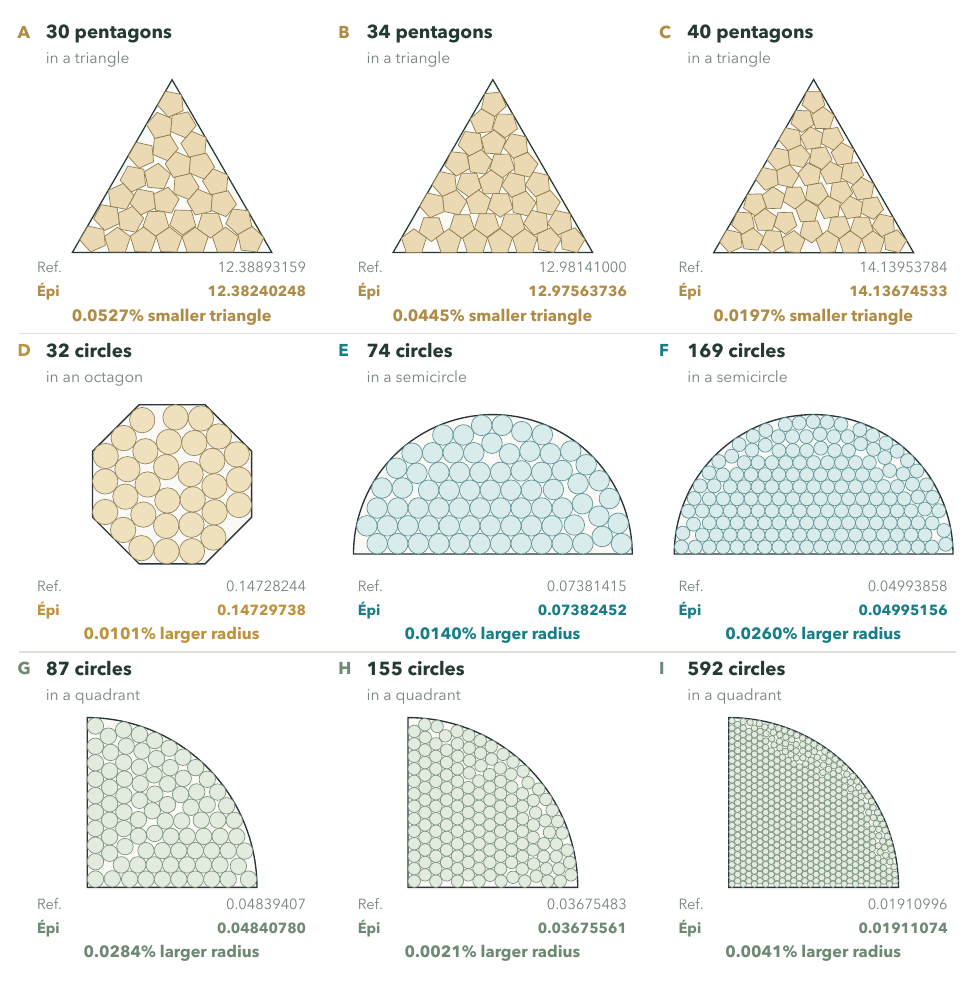}
\caption{Nine final record constructions. The three pentagon arrangements use unit-side pentagons; circle packings use a unit-radius semicircle or quadrant, or a unit-circumradius octagon. Percentages compare the final artifact with its archived reference. Triangle side length decreases; circle radius increases.}
\label{fig:packing}
\end{figure}

Figure~\ref{fig:packing} shows nine retained geometric constructions. For pentagons, the objective reduces the size of the enclosing triangle. The circle-packing tasks increase the common radius inside a fixed container \citep{packomania}. Each panel is drawn directly from the released coordinates.

\subsection{Separating the Cohn--Elkies Bound from Packing Density}
\label{sec:ce}
The Cohn--Elkies method bounds sphere-packing density from above by optimizing an auxiliary function. We ask whether its best possible bound can equal the actual packing density. A strict separation would establish a limitation of the method itself.

Let $\rho_d$ be the optimal sphere-packing center density in dimension $d$, and let $C_d$ be the best upper bound attainable by the Cohn--Elkies linear program in the same normalization. Then $\rho_d\leq C_d$. To prove that the method is not sharp, it is enough to find a packing upper bound $U_d$ and a lower bound $L_d$ on the linear program's optimum such that
\begin{equation}
 \rho_d\ \leq\ U_d\ <\ L_d\ \leq\ C_d.
 \label{eq:cegap}
\end{equation}
The middle inequality creates a separation between the true packing problem and the best answer available from this method.

\citet{li} obtains such lower bounds through a discrete reduction of the infinite-dimensional linear program. The reduction yields a finite optimization problem whose dual witnesses can be checked. \epi\ sets new records in dimensions 9, 10, 11, and 13, retaining one final dual vector for each. Table~\ref{tab:ce} reports the resulting lower bounds on $C_d$.

\begin{table}[htb]
\caption{New lower bounds on the optimal Cohn--Elkies bound in dimensions 9, 10, 11, and 13, rounded downward. The packing upper bounds are converted from Table 1.1 of \citet{cohn} to center density and rounded upward. All values use center-density normalization.}
\label{tab:ce}
\small
\begin{tabularx}{\linewidth}{C C C C}
\toprule\headrow
\textbf{Dimension} & \textbf{Earlier dual bound} & \textbf{\epi\ dual bound $L_d$} & \textbf{Packing upper bound $U_d$}\\\midrule
9 & $1/20$ & $0.058044329039$ & \\
10 & $1/24$ & $0.056433853144$ & $0.05623600$\\
11 & $1/24$ & $0.057731808202$ & $0.05664550$\\
13 & $1/28$ & $0.066440082613$ & \\\bottomrule
\end{tabularx}
\end{table}

\begin{figure}[htb]
\centering\includegraphics[width=\linewidth]{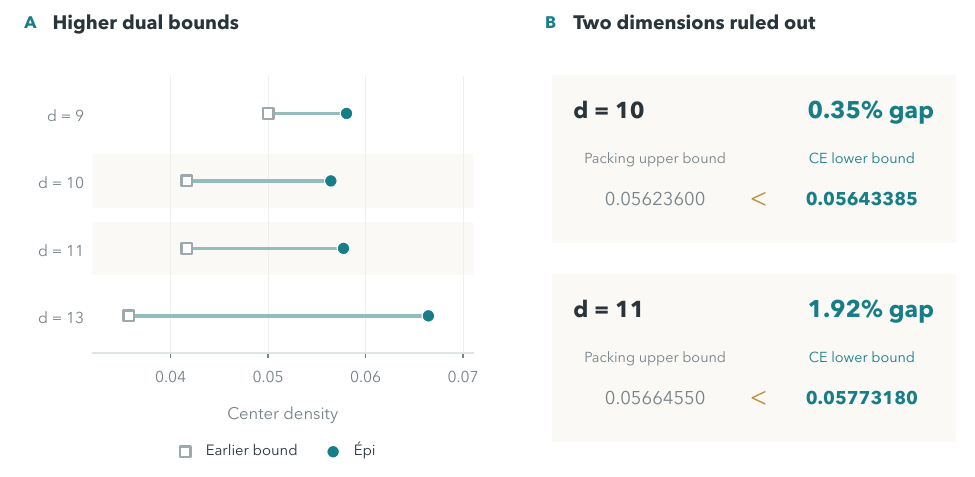}
\caption{Left: paired markers compare the earlier and improved dual lower bounds on the Cohn--Elkies optimum in each dimension. Right: in dimensions 10 and 11, the new lower bounds exceed known upper bounds on actual packing density. The displayed values use center density, and the relative gaps are $L_d/U_d-1$.}
\label{fig:ce}
\end{figure}

In dimensions 10 and 11, the new lower bounds cross the independently established packing upper bounds of \citet{cohn}. Equation~\eqref{eq:cegap} therefore holds in both dimensions. The relative separations, $L_d/U_d-1$, are approximately $0.35\%$ and $1.92\%$, respectively.

The separation identifies a structural limit of the Cohn--Elkies formulation in dimensions 10 and 11. Even exact optimization of this formulation cannot recover the true packing density; a sharp bound requires additional constraints or a different bounding method.

\subsection{Narrowing the Interval for the \texorpdfstring{$n$}{n}-Queens Constant}
The number $Q(n)$ of placements of $n$ non-attacking queens on an $n\times n$ board has an asymptotic form governed by a constant $\alpha$,
\begin{equation}
 Q(n)=\left(n e^{-\alpha+o(1)}\right)^n.
\end{equation}
Bounds on $\alpha$ can be obtained from finite convex optimization problems. \citet{nqueens} develop a scalable Newton method for these formulations and report the interval
\[
 [1.944000752,\;1.944001082].
\]

\begin{figure}[htb]
\centering\includegraphics[width=\linewidth]{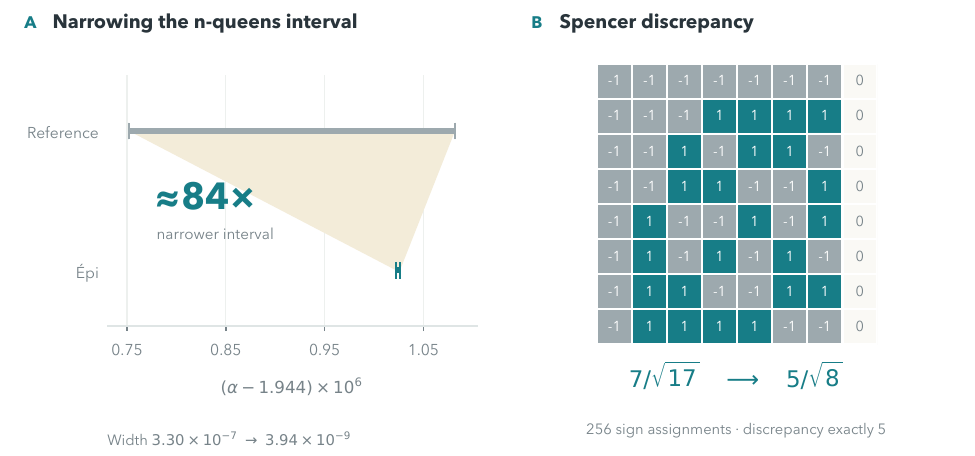}
\caption{Two new mathematical records. Left: the interval for the $n$-queens constant is approximately 84 times narrower. Right: an explicit $8\times8$ matrix with discrepancy five gives the Spencer lower bound $5/\sqrt8$. The matrix entries are $-1$, $0$, and $1$.}
\label{fig:mathprogress}
\end{figure}

\epi\ refines the lower and upper bounds separately. Rational feasibility checks and directed rounding give
\begin{equation}
 1.9440010223327
 \ \leq\ \alpha\ \leq\
 1.9440010262694.
 \label{eq:nq}
\end{equation}
The interval width decreases from $3.30\times10^{-7}$ to approximately $3.94\times10^{-9}$, a factor of approximately $84$. Figure~\ref{fig:mathprogress} places the two intervals on a common axis.

The two formulations have different computational costs, so Épi uses a higher resolution for the lower bound. Each endpoint has a retained witness that can be checked separately. Appendix~\ref{app:math} gives the resolutions.

\subsection{Optimizing a Formulation: MAX-4-CUT}
\label{sec:maxcut}
Épi starts from the MAX-4-CUT gadget framework of \citet{maxcut} and optimizes its edge weights. With the gadget structure and relevant completeness assignments fixed, the competing colorings have objectives that are linear in those weights.

This allows the agent to normalize one side of the hardness ratio and optimize the other through a linear program. Coloring constraints enforce the required properties of the gadget.

The reference ratio follows from the gadget parameters in Lemma C.2 and the reduction in Theorem C.2 of \citet{maxcut}. The final integer weighted gadget changes it from
\begin{equation}
 \frac{195581}{198140}\approx0.987085
 \quad\text{to}\quad
 \frac{84931}{86101}\approx0.986411.
\end{equation}
The released weighted edge list and finite checks recover this ratio exactly. Substituting the new gadget into the published reduction gives the improved hardness bound. This formulation provides a reusable optimization procedure: researchers can fix the gadget structure, optimize its weights subject to explicit coloring constraints, and verify the resulting ratio exactly.

\subsection{A Finite Construction for the Spencer Constant}
The square-matrix discrepancy problem asks how well signs can balance the rows of a real matrix with entries in $[-1,1]$. \epi\ found an $8\times8$ matrix whose discrepancy is exactly five. Exhausting its 256 sign assignments gives row-maximum norms of five or seven, so the minimum is five. The matrix therefore raises the record lower bound on the universal constant from $7/\sqrt{17}$ to $5/\sqrt8$ \citep{registry}.

Figure~\ref{fig:mathprogress} shows the entire matrix. Its entries lie in $\{-1,0,1\}$, within the allowed range $[-1,1]$, and the claim can be checked by enumerating its sign assignments.

\subsection{Certifying Near-Optimal Water-Network Design}
The waterund36 instance in MINLPLib minimizes the cost of a water-using network under nonlinear flow and concentration constraints \citep{minlplib}. A feasible network costing approximately $662.807$ was already known, but the listed global lower bound was only about $655.544$.

Épi tightens the lower bound to within approximately $10^{-5}$ of that same feasible value, closing more than $99.999\%$ of the MINLPLib gap. This certifies the near-optimality of the existing network. The full-precision comparison appears in Appendix~\ref{app:water}.

To establish the bound, the agent constructs a relaxation containing every feasible network. Exact rational inequalities then rule out objective values below the stated cutoff. The certificate therefore applies to the original nonlinear problem. Appendix~\ref{app:water} explains this argument and relates the instance to earlier optimization work.

\begin{figure}[htb]
\centering\includegraphics[width=\linewidth]{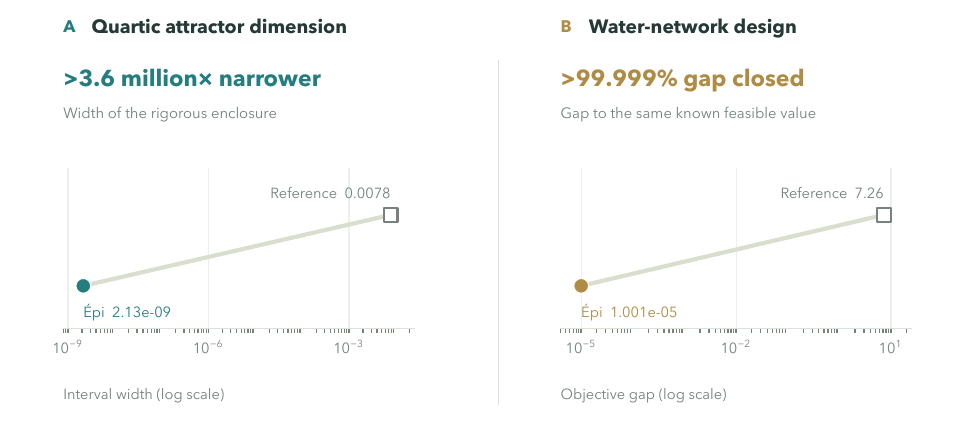}
\caption{Two improvements in certified uncertainty. The horizontal axes show interval or bound-gap width on logarithmic scales. Left: the rigorous enclosure for the quartic Feigenbaum attractor dimension becomes over 3.6 million times narrower. Right: the waterund36 gap to the same known feasible value decreases from about $7.263$ to $1.001\times10^{-5}$.}
\label{fig:certifiedgaps}
\end{figure}

\subsection{From Numerical Optimization to Proof: Tuza's Conjecture}
\label{sec:tuza}
For a finite simple graph $G$, let $\nu_{\triangle}(G)$ be the maximum number of pairwise edge-disjoint triangles and $\tau_{\triangle}(G)$ the minimum size of a \emph{triangle cover}: a set of edges meeting every triangle in $G$. Equivalently, deleting a triangle cover makes the entire graph triangle-free. Tuza's conjecture, proposed in 1981 \citep{tuza-threshold}, asserts
\begin{equation}
 \tau_{\triangle}(G)\leq 2\nu_{\triangle}(G).
 \label{eq:tuza}
\end{equation}
Thus, if $G$ contains no $k+1$ pairwise edge-disjoint triangles, deleting at most $2k$ edges suffices to eliminate every triangle in $G$.

\paragraph{Packing and covering}
K\H{o}nig's theorem equates the maximum matching size and minimum vertex-cover size in a bipartite graph \citep{konig1931}. Tuza's conjecture concerns a matching--cover relation in the \emph{triangle hypergraph}, whose vertices are the edges of $G$ and whose hyperedges are its triangles. Taking all edges of a maximum triangle packing gives the elementary bound $\tau_{\triangle}\leq3\nu_{\triangle}$: any uncovered triangle could be added to the packing. The conjectured factor of two is attained by $K_4$, where $\nu_{\triangle}=1$ and $\tau_{\triangle}=2$.

\paragraph{Split graphs and prior results}
A split graph has a partition $V(G)=C\sqcup I$, where $C$ induces a clique and $I$ is independent. Known cases include threshold graphs, whose independent-side neighborhoods are nested by inclusion \citep{tuza-threshold}, and $K_8$-free chordal graphs, which include $K_8$-free split graphs \citep{tuza-treewidth}. Our eight-vertex clique part puts the family beyond the latter theorem's $K_8$-free hypothesis.

\paragraph{Neighborhood types and the extension}
For $x\in I$, let $S_x=N_G(x)\cap C$. Since $C$ is a clique, every pair $u,v\in S_x$ forms a triangle $xuv$. Thus $x$ is \emph{triangle-active} exactly when $|S_x|\geq2$; vertices with identical $S_x$ share one \emph{neighborhood type}. \citet{tuza-zeng} proved the conjecture for $|C|=8$ and at most two types. \epi\ extends this to three types with unrestricted overlaps. Their triangles can compete for the same clique edges, requiring joint covers and packings.

\begin{theorem}[Three neighborhood types]
\label{thm:tuza}
Let $G$ be a finite simple graph with a specified partition $V(G)=C\sqcup I$, where $G[C]=K_8$ and $I$ is independent. Suppose that the sets $N_G(x)\cap C$, over vertices $x\in I$ with at least two neighbors in $C$, take at most three distinct values. Then $\tau_{\triangle}(G)\leq 2\nu_{\triangle}(G)$.
\end{theorem}

Up to clique relabeling and type permutation, there are 872 three-type profiles: 35 have nested neighborhoods and 277 further profiles satisfy Zeng's maximum-cut criterion. \epi\ constructs exact cover and packing certificates for the remaining 560 profiles, using 18,444 multiplicity boxes. An independent checker verifies the witnesses and complete profile enumeration. Zeng's truncation lemma extends the checks to arbitrary multiplicities, establishing the result for an infinite graph family (Appendix~\ref{app:tuza}).

The proof separates the finite certificate checks from the argument extending them to arbitrary multiplicities, making explicit how the computational evidence establishes the infinite-family statement.

\FloatBarrier
\section{Épi-turbo: Fast, Low-cost Algorithm Design}
\label{sec:turbo}
\subsection{Purpose and Design}
Algorithmic coding extends Épi's research process to the development of executable solvers. Given a problem specification and a fixed evaluator, the agent designs, implements and tests an algorithm whose solution quality and runtime are measured on new instances. Épi-turbo is a lightweight configuration for these tasks, designed to develop effective algorithms within a short campaign and a small model budget.

\paragraph{Shared foundation}
Épi-turbo keeps the parts of Épi that make a result reliable. The verifier checks each candidate under a fixed ruler, and Épi keeps the research state across sessions. Branches keep each method with its code and evidence, so the agent can return to an earlier version. Memory records keep lessons and dead ends for later sessions. Épi-turbo changes how the agent uses its model budget and when the campaign stops.

\paragraph{Instructions for algorithm tasks}
The task instructions tell the agent how the benchmark measures a program. They give the complete score formula, which includes the quality term and the runtime term. They also tell the agent that the final score comes from instances that the agent does not see. When a program uses a large part of its time limit, Épi-turbo tells the agent. Thus the agent can find a good balance between solution quality and runtime. The instructions do not contain methods or hints for a specific problem.

\paragraph{Model calls for algorithm design}
Each model call has a cost, and a short task has a small number of calls. Épi-turbo tells the agent that every call costs money, which includes a check on an active job. The agent does short experiments with a time limit in its own session. It sends an experiment to the execution infrastructure only when the experiment needs more time than a session gives. It examines active jobs only when a result can change its next decision. As a result, most of the budget goes to the design and test of algorithms.

\paragraph{Continuous sessions}
Épi-turbo runs each campaign as one continuous session if possible. The session continues the best method so far and does not restart from an empty context. Before a session ends, the agent records its state, its best candidate and its next steps. The next session reads this record and continues the work. Épi-turbo stops the campaign when more sessions do not improve the best verified score. Thus the agent does not use more budget on a task where it finds no improvement.

\paragraph{Budget control}
Épi-turbo reserves the maximum cost of a model call before the call starts. It also keeps a small part of the budget for the end of the campaign. Thus the cost of a campaign is never more than its budget, and the final steps always have funds. The reservation uses the price limits of the model, so the agent can use almost all of its budget.

\paragraph{Optional components}
Independent review and extra validation are optional in Épi-turbo. On short tasks, their cost was larger than their benefit in our tests. Both components were off in the runs below.

\subsection{Benchmark and Protocol}
We use FrontierOR, a benchmark of 180 operations-research tasks from the literature \citep{frontieror}. Each task asks for a standalone program that solves new instances of one problem class. The program must not call an external optimization solver. The baseline is Codex \citep{codex}, a widely used coding agent, with GPT-6-Astra as its model. Épi-turbo uses the same model with the same reasoning effort.

We selected 30 tasks that were new to both systems. We took them in the official hardness order of the benchmark and excluded tasks with unreliable reference values. Of these tasks, 13 are in the hard split, and 17 have hardness ranks 51 to 70. For each task, each system had a model budget of \$20. We did one run for each task and system, so we do not report statistical significance.

The evaluator first runs each program on two development instances. The agent sees these results during the campaign. After the campaign, the best development program runs once on four held-out instances. Each held-out run has a time limit of one hour. The benchmark score has a quality term and a speed term.

The quality term decreases as the gap to the reference value increases. The speed term compares the runtime of the program with the runtime of the reference solver. It counts only when the solution is within 1\% of the reference value. Outside this 1\% band, the score is the quality term alone. The reference values come from a commercial solver, and some of them are weak. Thus we also compare the two systems on each instance directly, with their objective values only.

\subsection{Results}
Table~\ref{tab:turbo} gives the main results. Épi-turbo gets the higher total score, 53.96 against 51.21. It has the higher score on 18 tasks, and Codex has the higher score on 12. On 10 of these tasks, the difference is less than 0.01. Épi-turbo uses 64\% of the model cost of Codex. Its median research time per task is approximately 22 minutes, and the median for Codex is 37 minutes.

\begin{table}[htb]
\caption{Épi-turbo and Codex on 30 FrontierOR tasks that were new to both systems. Both systems use GPT-6-Astra and a model budget of \$20 per task. All scores are for the held-out instances. The instance counts compare the objective values of the two programs directly.}
\label{tab:turbo}
\small
\begin{tabularx}{\linewidth}{L >{\centering\arraybackslash}m{2.6cm} >{\centering\arraybackslash}m{2.6cm}}
\toprule\headrow
\textbf{Measure} & \textbf{Épi-turbo} & \textbf{Codex}\\\midrule
Total benchmark score & $53.96$ & $51.21$\\
Tasks with the higher score & $18$ & $12$\\
Mean quality term & $1.030$ & $1.032$\\
Mean speed term & $0.784$ & $0.675$\\
Instances with the better solution (119 comparable, 40 ties) & $18$ & $61$\\
Instances more than 1\% worse than the other system & $29$ & $10$\\
Infeasible solutions (120 held-out runs) & $1$ & $0$\\
Model cost per task, mean & \$7.30 & \$11.35\\
Research time per task, median & 22 min & 37 min\\\bottomrule
\end{tabularx}
\end{table}

\begin{figure}[htb]
\centering\includegraphics[width=\linewidth]{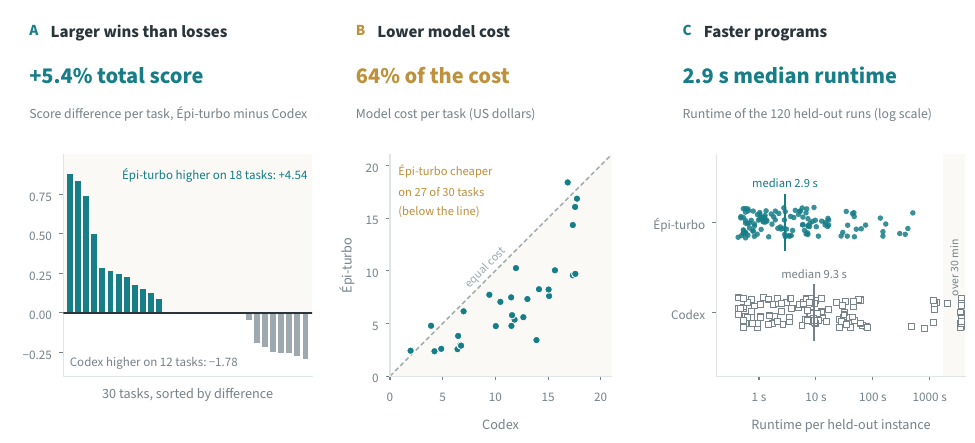}
\caption{Épi-turbo and Codex on 30 FrontierOR tasks. A: The score difference for each task, in order of size. Épi-turbo has the higher score on 18 tasks, and its gains are larger than its losses. B: The model cost for each task. Each point below the dashed line is a task where Épi-turbo costs less. C: The runtime of each held-out run on a logarithmic scale, with the medians as vertical bars.}
\label{fig:turbo}
\end{figure}

Figure~\ref{fig:turbo} shows the results for each task. The gains of Épi-turbo are larger than its losses. On the 18 tasks where Épi-turbo is higher, the total gain is 4.54. On the other 12 tasks, the total loss is 1.79. Épi-turbo also costs less on 27 of the 30 tasks.

Table~\ref{tab:turbo-groups} splits the tasks into two groups. On the 13 tasks of the hard split, Épi-turbo has the higher score on 9 tasks, and its total score is 10\% higher. On the 17 tasks with ranks 51 to 70, the two systems are close. On 15 of the 30 tasks, Épi-turbo has both the higher score and the lower model cost.

\begin{table}[htb]
\caption{Results by task group. The hard split is the official set of the 50 hardest FrontierOR tasks. Each pair of values gives Épi-turbo first and Codex second.}
\label{tab:turbo-groups}
\small
\begin{tabularx}{\linewidth}{L >{\centering\arraybackslash}m{1.2cm} C C C}
\toprule\headrow
\textbf{Group} & \textbf{Tasks} & \textbf{Total score} & \textbf{Higher score} & \textbf{Model cost}\\\midrule
Hard split & 13 & $22.33$ / $20.30$ & $9$ / $4$ & \$95 / \$152\\
Ranks 51--70 & 17 & $31.64$ / $30.91$ & $9$ / $8$ & \$124 / \$188\\
All tasks & 30 & $53.96$ / $51.21$ & $18$ / $12$ & \$219 / \$341\\\bottomrule
\end{tabularx}
\end{table}

The higher score comes mainly from the speed term. The programs of Épi-turbo usually stop when their search finds no better solution. On the held-out instances, the median runtime of an Épi-turbo program is 2.9 seconds, and the median for Codex is 9.3 seconds. No held-out run of Épi-turbo used more than 30 minutes. Codex used more than 30 minutes on 17 of the 120 held-out runs, in 7 tasks. A long run can improve the solution, but it decreases the speed term.

\subsection{Selected Cases}
Table~\ref{tab:turbo-wins} lists the ten algorithmic coding tasks with the largest score difference in favor of Épi-turbo. The tasks span vehicle routing, cooperative crane scheduling, production planning, facility location, clustering and graph optimization. Each row evaluates a generated program on unseen instances of its problem class. This breadth makes the connection to the mathematical campaigns explicit: Épi's research process supports both constructing mathematical artifacts and developing algorithms with measurable gains in practical optimization settings. The four cases below show how solution quality and execution speed contribute to these gains.

\paragraph{Clique interdiction: better solutions and faster programs (furini2021)}
The Épi-turbo program solves each held-out instance in less than 17 seconds. On one instance, its solution has an objective value of 8, and the Codex solution has 20. The Codex program also runs for 20 to 60 minutes on three of the four instances. Thus Épi-turbo has both the higher quality term and the higher speed term.

\paragraph{Production planning: reaching the 1\% band (delage2022)}
Épi-turbo has the better solution on three instances and the same solution on the fourth. On one instance, the Codex solution is 1.19\% above the reference value, and the Épi-turbo solution is 0.89\% above it. Thus only Épi-turbo gets the speed term on that instance. The Épi-turbo programs are also 2 to 7 times faster.

\paragraph{Maximum clique: matching the reference (furini2019)}
Épi-turbo matches the reference value on all four instances. The Codex solution is 5\% worse on one instance. On another instance, the Codex program needs 34 seconds, which is more time than the reference solver needs. Thus Codex loses the speed term on that instance, and Épi-turbo keeps it.

\paragraph{Vehicle routing: faster execution (he2020)}
Both systems beat the reference values by 5\% to 17\%. The Codex solutions are slightly better on three instances, by at most 2.6\%. But the Codex program uses the full hour on three instances, and the Épi-turbo program needs at most 36 seconds. Here the speed term decides the result.

\begin{table}[htb]
\caption{The ten FrontierOR coding tasks with the largest score difference in favor of Épi-turbo, sorted by the difference. Task names identify the optimization problem solved by each generated program; benchmark IDs appear below them. Scores, quality and speed are measured on held-out instances. Each pair gives Épi-turbo first and Codex second.}
\label{tab:turbo-wins}
\small
\begin{tabularx}{\linewidth}{>{\raggedright\arraybackslash}m{4.1cm} C C C C}
\toprule\headrow
\textbf{Task} & \textbf{Score} & \textbf{Quality term} & \textbf{Speed term} & \textbf{Model cost}\\\midrule
Hyperplane clustering\newline{\footnotesize amaldi2013} & $2.162$ / $1.283$ & $1.163$ / $1.282$ & $0.999$ / $0.001$ & \$2.59 / \$4.86\\
Clique interdiction\newline{\footnotesize furini2021} & $2.090$ / $1.256$ & $1.091$ / $0.841$ & $0.999$ / $0.415$ & \$3.82 / \$6.47\\
Vehicle routing with node costs\newline{\footnotesize he2020} & $2.094$ / $1.353$ & $1.098$ / $1.108$ & $0.995$ / $0.245$ & \$3.42 / \$13.90\\
Maximum clique\newline{\footnotesize furini2019} & $1.987$ / $1.487$ & $1.000$ / $0.988$ & $0.987$ / $0.500$ & \$5.78 / \$11.58\\
Cooperative crane scheduling\newline{\footnotesize kress2019} & $1.853$ / $1.572$ & $1.103$ / $1.074$ & $0.750$ / $0.498$ & \$4.78 / \$3.90\\
Process configuration and production planning\newline{\footnotesize delage2022} & $2.023$ / $1.759$ & $1.024$ / $1.012$ & $0.999$ / $0.747$ & \$7.32 / \$13.04\\
Routing with time-window assignment\newline{\footnotesize spliet2015} & $2.012$ / $1.765$ & $1.012$ / $1.016$ & $1.000$ / $0.750$ & \$7.48 / \$11.50\\
Graph coloring\newline{\footnotesize gualandi2012} & $2.086$ / $1.860$ & $1.117$ / $1.128$ & $0.969$ / $0.731$ & \$4.77 / \$11.52\\
Set partitioning\newline{\footnotesize zhang2025} & $1.558$ / $1.379$ & $1.021$ / $0.951$ & $0.536$ / $0.428$ & \$18.38 / \$16.87\\
Facility location\newline{\footnotesize fischetti2017} & $1.998$ / $1.847$ & $1.000$ / $1.000$ & $0.998$ / $0.847$ & \$2.41 / \$1.96\\\bottomrule
\end{tabularx}
\end{table}

\FloatBarrier
\section{Related work}
\label{sec:related}
Autonomous research systems differ in how they organize the search for a useful next experiment. The AI Scientist-v2 explores experimental directions through agentic tree search \citep{ai-scientist}. Co-Scientist uses model-generated critique to refine hypotheses around a research goal \citep{co-scientist}, and InternAgent-1.5 coordinates generation with verification across computational and empirical tasks \citep{internagent}. These systems share Épi's need to revise a research direction in response to feedback. AIRA$_2$ addresses a related execution challenge through asynchronous experiments and a consistent evaluation protocol \citep{aira2}. In Épi, the verifier stays fixed within a campaign as the agent changes its approach.

Continuing research also depends on how previous work is represented. Kosmos uses a shared world model to connect literature search and data analysis \citep{kosmos}. ScienceFlow preserves executable research states that can be resumed or revisited \citep{scienceflow}. Knowledge can also pass between investigations through reports, as in AgentRxiv \citep{agentrxiv}, or through AutoKD's persistent graph of accepted findings \citep{autokd}. Épi retains both the experimental state needed to resume a campaign and lessons intended for reuse on another problem. Branches preserve alternative histories within that campaign.

For tasks with executable evaluation, FunSearch and AlphaEvolve guide program generation using measured performance \citep{funsearch,alphaevolve}. SimpleTES scales search through independent trajectories and reuse of evaluated histories \citep{simpletes}. Dream-RSI turns recorded discovery trees into replay environments for evaluating and revising exploration policies before deploying them in further searches \citep{dream-rsi}. TTT-Discover adapts model weights at test time \citep{ttt-discover}. Épi keeps model weights fixed and uses stored lessons and branch histories to guide its next experiment. Released Hyra constructions offer starting points for further refinement \citep{hyra}. Épi's investigations can also extend beyond refinement to finding a certificate or proving a statement, as in the Tuza case study.

\Needspace{8\baselineskip}
\section{Autonomous Research for Human Understanding}
\label{sec:human-understanding}
Our goal with Épi is to make autonomous research contribute to human understanding: researchers should be able to explain why a result holds, identify what it reveals about the problem, and use that knowledge to guide further work. Achieving this goal requires connecting computational artifacts to explicit formulations, mathematical arguments, and the scientific questions they resolve. Épi builds on existing theory and retains the implementations, experiments, and assessments through which its methods develop. The case studies illustrate how this process can produce knowledge that researchers can use: the Cohn--Elkies separation identifies a structural limit of an established method; the MAX-4-CUT formulation provides a reusable procedure for optimizing gadget weights; and the Tuza proof makes explicit how finite certificates establish a statement about an infinite graph family. These examples express our broader aim for self-improvement: an agent should accumulate both the ability to obtain better results and knowledge that helps people understand and extend those results.

\section{Conclusion}
Épi is an autonomous research agent that sustains investigations across model sessions and improves its methods through memory, branching and independent review. Its campaigns produced 120 record-setting results and a computer-assisted proof of a new case of Tuza's conjecture. The released constructions and certificates provide concrete, checkable evidence for these advances. Épi-turbo extends this research process to algorithmic coding across routing, scheduling, planning and graph optimization. On 30 FrontierOR tasks, it achieves a 5.4\% higher total benchmark score than Codex with the same model while using 64\% of its model cost.

Épi couples method improvement with a persistent record of implementations, experiments, and review. This record supports inspection of the research process, while explicit constructions and proof arguments make the resulting findings available for further scientific work. Together, these evaluations demonstrate the applicability of persistent, feedback-driven method development to mathematical discovery and efficient algorithm design.

\section*{Acknowledgements}
{\raggedright
We thank \mbox{Prof.~Henry Cohn}, \mbox{Prof.~David Galvin}, \mbox{Dr.~Edward Hirst}, \mbox{Dr.~Adam Onus}, and \mbox{Prof.~Erich Friedman} for their valuable feedback.\par}

\FloatBarrier
\setlength{\bibsep}{0.2pt plus 0.2ex}
\begingroup
\raggedright
\renewcommand{\bibfont}{\linespread{1}\fontsize{9}{10.5}\selectfont}
\bibliographystyle{plainnat}
\bibliography{references}
\endgroup
\clearpage
\appendix
\section{Mathematical Results and Supporting Evidence}
\label{app:math}
This appendix records the twelve numerical result families and the Tuza theorem presented in the report. Numerical entries state the new bound, its previous record, and the final artifact that supports it. The Tuza entry explains the finite reduction and certificates behind Theorem~\ref{thm:tuza}. The four Cohn--Elkies dimensions, five $B_h[g]$ coefficients and two quantum channels each count as separate parameter settings in the 20 numerical records.

\subsection{Relevant Variables of Boolean Functions}
The record bound on Wellens' quantity $D_\infty$ decreases from $4.3935$ to $4.3654674227$ \citep{wellens}. This quantity controls the number of relevant variables relative to Boolean-function degree. The finite computation retains an exact infeasibility certificate and a rational feasibility witness for each degree from 1 to 48. The published formulation combines these per-degree certificates into the bound on $D_\infty$.

\subsection{Cohn--Elkies Linear Programming}
The final dual vectors establish the lower bounds in Table~\ref{tab:ce}, using Li's discrete reduction \citep{li}. Each dimension has a retained vector and an exact bound expression. The comparison with three-point packing upper bounds \citep{cohn} establishes the separation in dimensions 10 and 11 described in Section~\ref{sec:ce}. These values bound the optimum of the Cohn--Elkies program from below.

\subsection{Spencer Discrepancy Constant}
For an $n\times n$ real matrix $A$ with entries of magnitude at most one, define
\[
 \operatorname{disc}(A)=\min_{\varepsilon\in\{-1,1\}^n}\|A\varepsilon\|_\infty.
\]
An explicit $8\times8$ matrix in the collection has discrepancy five. Exhausting all 256 sign assignments verifies this value. It yields the lower bound $5/\sqrt8\approx1.767767$ on the universal constant in the square-matrix discrepancy problem, improving the reference $7/\sqrt{17}\approx1.697749$ \citep{registry}. The final artifact is the matrix itself.

\subsection{Dimer Entropy in Three Dimensions}
The record upper bound on the per-site entropy of perfect matchings of the cubic lattice decreases from $0.452130$ to $0.452010$, extending the transfer-operator computation of \citet{dimer}. The retained positive integer vectors satisfy the transfer-operator inequalities. Evaluating the resulting logarithm gives approximately $0.452009597554483$. The stated bound is rounded upward.

\subsection{The \texorpdfstring{$n$}{n}-Queens Constant}
The final lower and upper witnesses have resolutions 65,536 and 4,096, respectively. Rational feasibility checks and directed rounding give Equation~\eqref{eq:nq}. The interval is approximately 84 times narrower than the comparator from \citet{nqueens}. Both witnesses are retained, so the two sides can be checked separately.

\subsection{Ramsey Multiplicity \texorpdfstring{$c_{4,5}$}{c4,5}}
A graph on 1,024 vertices changes the record upper bound from
\[
 \frac{2129191}{2^{28}}
 \quad\text{to}\quad
 \frac{4427841832181}{2^{49}}.
\]
This is the asymmetric multiplicity problem involving a clique of order four in one color and a clique of order five in the other, following \citet{ramsey}. The exact reference value is computed from the published 128-vertex graph in the accompanying dataset \citep{ramsey-data}. The final graph preserves the construction. A separate exact enumeration checks the graph and its complement.

\subsection{MAX-4-CUT Hardness}
The gadget in Section~\ref{sec:maxcut} gives the exact ratio $84931/86101$, compared with $195581/198140$ \citep{maxcut}. Integer edge weights make the gadget's finite objectives recoverable without relying on floating-point optimizer output. The hardness statement uses the original reduction, with the improved weighted gadget substituted into it.

\subsection{Polyomino Growth: Klarner's Constant}
The number of fixed polyominoes grows exponentially at rate $\lambda$. Extending the neighborhood recurrence method of \citet{klarner}, \epi\ improves the published upper bound $4.5238$ to
\[
 \lambda\leq\frac{429569}{100000}=4.29569.
\]
The certificate covers 1,186 neighborhood types. A separate checker reconstructs the geometric rules, checks the recurrence dependencies and verifies the rational supersolution. Verifying the supersolution against every recurrence constraint establishes the bound for polyominoes of all sizes.

\subsection{Additive Combinatorics: \texorpdfstring{$B_h[g]$}{Bh[g]} Sets}
A $B_h[g]$ set has at most $g$ unordered representations of any integer as a sum of $h$ elements. For $g\geq2$, the framework of \citet{bhg} bounds the size of such a set $A\subseteq[N]$ by
\[
 |A|\leq(1+o_N(1))(C_hgN)^{1/h}.
\]
\epi\ improves all five coefficients below. The published column follows Timmons' Table~1. The new values are rounded upward.
\begin{table}[H]
\centering\small
\caption{Improved upper-bound coefficients for $B_h[g]$ sets.}
\label{tab:bhg}
\begin{tabularx}{\linewidth}{C C C}
\toprule\headrow
$h$ & \textbf{Published coefficient} & \textbf{\epi\ coefficient}\\\midrule
3 & $14.3$ & $14.287390218082$\\
4 & $71.49$ & $70.289552594834$\\
5 & $413.07$ & $408.104453996958$\\
6 & $2774.16$ & $2749.492485514635$\\
7 & $21294.74$ & $21149.189776156220$\\\bottomrule
\end{tabularx}
\end{table}
Each certificate supplies a rational cosine kernel. Interval arithmetic establishes its minimum on the full domain and propagates that bound to $C_h$. Earlier and new kernels are evaluated by the same frozen implementation.

\subsection{Quartic Feigenbaum Attractor Dimension}
\citet{feigenbaum} give a rigorous enclosure of the quartic attractor's Hausdorff dimension. Using their rigorous fixed-point enclosure, \epi\ proves
\[
 0.6425750638035370\leq\dim_H(A_4)\leq0.6425750659339700.
\]
The published interval has endpoints $0.6395131468772885$ and $0.6473156929016112$ (rounded outward). The new interval has width $2.130433\times10^{-9}$, more than $3.6$ million times narrower than the cited rigorous interval. A separate implementation checks the pressure inequalities throughout the domain.

\subsection{Quantum-Capacity Thresholds}
Following the permutation-invariant state formulation of \citet{qcap}, \epi\ finds 40-qubit input states with positive coherent information at
\[
 p_{\mathrm{XZ}}=\frac{1198803}{10000000}=0.1198803,
 \qquad
 p_{\mathrm{2P}}=\frac{238683}{2000000}=0.1193415.
\]
These improve the reported independent X--Z and Two-Pauli thresholds of approximately $0.118371$ and $0.118067$, respectively. Separate interval spectral calculations verify positive coherent information for both states, establishing positive quantum capacity at the stated noise parameters.

\subsection{Water-Network Design: Waterund36}
\label{app:water}
For the original continuous waterund36 minimization problem \citep{minlplib}, the certificate establishes
\[
 f^*\geq662.80702839.
\]
It excludes objective values at or below the exact rational cutoff
\[
 \frac{6628070283996671}{10^{13}}.
\]
The displayed lower bound is rounded downward. Compared with MINLPLib's lower bound $655.5441707$ and known feasible value $662.8070384$, the remaining gap is about $0.00001001$. The instance also appears in Castro's study of global optimization with mixed-integer relaxations \citep{castro}.

The proof rules out every network cheaper than the cutoff by deriving a contradiction from that cost assumption. It first lifts each feasible network into a bounded relaxation using flow and contaminant conservation. For a linear relaxation $Ax\leq b$ with box bounds $L\leq x\leq U$, any nonpositive row multiplier vector $y$ gives the valid lower support
\[
 c^Tx\geq y^Tb+\min_{L\leq z\leq U}(c-A^Ty)^Tz.
\]
Equality-row multipliers may have either sign. The box minimum is evaluated componentwise. Rational support inequalities and bound propagation eventually yield incompatible bounds under the cutoff assumption. Containment of the original feasible set in the relaxation then establishes the global lower bound. The released certificate supplies the model, witnesses and exact comparisons needed for this argument.

\subsection{Tuza's Conjecture: Proof for Three Neighborhood Types}
\label{app:tuza}
For the partition in Theorem~\ref{thm:tuza}, discard independent vertices with fewer than two clique neighbors: they occur in no triangle. The remaining vertices have at most three distinct neighborhoods in $C$. We call a triple of these neighborhoods, considered up to relabeling $C$ and permuting the three types, a neighborhood profile.

\paragraph{Finite reduction}
Zeng's multiplicity-truncation lemma \citep{tuza-zeng} preserves both $\tau_{\triangle}$ and $\nu_{\triangle}$ when each type is truncated to 28 copies. Every triangle uses a clique edge, so a packing uses at most 28 triangles and at most 28 independent vertices of any type. For covers, all 28 clique edges suffice. If a minimum cover of the truncated graph has fewer than 28 edges, some copy of every saturated type has no incident cover edge. All clique edges within that type's neighborhood must then be covered, which also covers triangles through any additional copies. Thus the finite multiplicity range suffices for the infinite statement.

\paragraph{Completeness and certificates}
An independent enumeration gives 872 profiles with three distinct active types. Of these, 35 have nested neighborhoods and fall under the threshold-graph theorem of \citet{tuza-threshold}. Another 277 satisfy Zeng's maximum-cut criterion \citep{tuza-zeng}. The cases with at most two types follow from Zeng's theorem. The remaining 560 profiles are covered by exact certificates containing 18,444 integer boxes in $\{0,\ldots,28\}^3$.

For a box with componentwise bounds $\ell\leq m\leq u$, the certificate gives a triangle cover $F_u$ meeting every triangle in the upper-multiplicity graph $G_u$ and an edge-disjoint triangle packing $P_\ell$ in the lower-multiplicity graph $G_\ell$, satisfying $|F_u|\leq2|P_\ell|$. The cover meets both the triangles inside $C$ and every triangle $xuv$ with $x\in I$ and $u,v\in N_G(x)\cap C$. Adding copies preserves the smaller graph as a subgraph. Hence
\[
 \tau_{\triangle}(G_m)\leq |F_u|\leq2|P_\ell|\leq2\nu_{\triangle}(G_m).
\]
A separate checker reconstructs the graphs, enumerates every triangle in each upper-multiplicity graph, checks that its stored cover meets every such triangle, and verifies that the stored packing consists of edge-disjoint triangles in the lower-multiplicity graph. It also checks the cardinality inequality and coverage of every integer multiplicity vector. All 560 profiles and 18,444 boxes passed. The certificates cover the remaining profiles, and truncation extends the conclusion to arbitrary multiplicities. This proves Theorem~\ref{thm:tuza}.

\FloatBarrier
\section{Additional Construction Results}
\label{app:construction}
For the 34-pentagon case in Section~\ref{sec:packing}, the enclosing triangle side length decreases from $12.98141$ to $12.97563735803487$. The released coordinates retain the construction at full precision.

Table~\ref{tab:nonpacking} records the three non-packing records in the construction collection. Each row compares the same task instance and objective convention. All objectives in this table are minimized.

\begin{table}[htb]
\caption{Non-packing construction results in the construction collection.}
\label{tab:nonpacking}\small
\begin{tabularx}{\linewidth}{>{\raggedright\arraybackslash}p{3.35cm} R R}
\toprule\headrow
\textbf{Problem} & \textbf{Archived reference} & \textbf{\epi}\\\midrule
Vehicle routing, XL-n1048-k237 & $380107$ & $380092$\\
Autocorrelation (signed maximum) & $1.4515718639$ & $1.4508066395$\\
Zhang--Zagier, $c_{82}$ & $0.2536331090$ & $0.2535744749$\\\bottomrule
\end{tabularx}
\end{table}

\paragraph{Vehicle routing}
The final construction contains 238 routes serving 1,047 customers with vehicle capacity 128. The objective uses the instance's rounded Euclidean distances \citep{cvrplib}. The route list determines customer coverage, capacity feasibility, and total length.

\paragraph{Autocorrelation}
The comparison uses the signed-maximum task and archived reference from \citet{einstein}. For signed step values $a_1,\ldots,a_n$, the objective is
\[
 J(a)=\frac{2n\left\lvert\max_j(a*a)_j\right\rvert}{(\sum_i a_i)^2},
\]
where $a*a$ is the discrete self-convolution. The absolute value is taken after the maximum. The maximum-absolute-value variant instead uses $\max_j\lvert(a*a)_j\rvert$ in the numerator; the two objectives can differ for signed constructions. Under the signed-maximum objective, the released construction improves the archived reference by approximately $7.65\times10^{-4}$.

\paragraph{Zhang--Zagier}
The comparison uses the archived reference from \citet{zhangzagier}. The $c_{82}$ construction has an accompanying Arb-based numerical certificate for its objective.

\end{document}